# Relative Value Biases in Large Language Models

**William M. Hayes (whayes2@binghamton.edu)**
Department of Psychology, Binghamton University, Binghamton, NY, USA

**Nicolas Yax (nicolas.yax@ens.psl.eu)**
**Stefano Palminteri (stefano.palminteri@ens.fr)**
Département d'Etudes Cognitives, Ecole Normale Supérieure, Paris, France

**Abstract**

Studies of reinforcement learning in humans and animals have demonstrated a preference for options that yielded relatively better outcomes in the past, even when those options are associated with lower absolute reward. The present study tested whether large language models would exhibit a similar bias. We had gpt-4-1106-preview (GPT-4 Turbo) and Llama-2-70B make repeated choices between pairs of options with the goal of maximizing payoffs. A complete record of previous outcomes was included in each prompt. Both models exhibited relative value decision biases similar to those observed in humans and animals. Making relative comparisons among outcomes more explicit magnified the bias, whereas prompting the models to estimate expected outcomes caused the bias to disappear. These results have implications for the potential mechanisms that contribute to context-dependent choice in human agents.

**Keywords:** artificial intelligence; large language models; machine psychology; value-based decision-making; reinforcement learning

## Introduction

Decision-making agents often face choices that are similar to ones they have encountered in the past. In many cases, the exact reward values of the options are unknown, but the outcomes of previous choices can be used to estimate expected payoffs and choose the option associated with greater reward. To make payoff-maximizing choices in a variety of contexts, agents need to possess accurate representations of the absolute values of previous outcomes.

However, evidence from multiple studies and multiple labs suggests that outcomes may be represented on a relative or partially relative scale (e.g., Bavard et al., 2018; Hayes & Wedell, 2023; Klein, Ullsperger, & Jocham, 2017; Palminteri et al., 2015). Options are assessed by how they compare to other options in the same choice set rather than an absolute standard (Vlaev et al., 2011). For example, an option associated with moderate rewards can be evaluated positively or negatively depending on context: If paired with a low-reward option, it may be evaluated positively, but if paired with a high-reward option, it may be evaluated negatively. In repeated choice tasks, relative valuation can facilitate learning by enhancing the differences between options that belong to a given context. At the same time, relative values cannot always be extrapolated to new contexts: An option that was best in one setting may not be best in another. Relying on relative values learned in one context to make decisions in another context may lead to suboptimal preferences (Bavard, Rustichini, & Palminteri, 2021). For example, an option might be chosen simply because it was better than the alternatives in its original context, even if that is no longer the case. Suboptimal preferences induced by relative valuation are frequently observed in human reinforcement learning experiments (for a review, see Palminteri & Lebreton, 2021), and recent evidence suggests that they generalize across cultures and socioeconomic backgrounds (Anlló et al., 2023). Relative value biases have also been observed in some animal species, including rodents (Flaherty, 1996), birds (Pompilio & Kacelnik, 2010), and insects (Solvi et al., 2022).

Given the pervasiveness of relative value biases in human and nonhuman species, a natural question is whether similar biases would be observed in artificially intelligent agents. The recent development of large language models (LLMs) that exhibit human-like abilities in a variety of language-based tasks makes this an opportune time to investigate such questions. LLMs are remarkably successful at *in-context learning*: they can figure out how to perform novel tasks using only the information contained in the prompt, without fine-tuning the weights (Brown et al., 2020; Kojima et al., 2022). In the burgeoning field of "machine psychology," researchers have begun treating LLMs as participants in psychological experiments and comparing the models' behavior in various tasks to human behavior (Hagendorff, 2023). The machine psychology approach has been successful at revealing similarities and differences in the ways that humans and LLMs make decisions in multiple domains (e.g., Aher, Arriaga, & Kalai, 2023; Binz & Schulz, 2023; Coda-Forno et al., 2023; Horton, 2023; Yax, Anlló, & Palminteri, 2023). Our goal was to extend this approach to the study of relative value biases in decision making.

To this end, we adapted a bandit task from a prior study on human reinforcement learning (Hayes & Wedell, 2023). The task is comprised of two parts: a learning phase and a transfer test. The learning phase consists of repeated choices between options presented in fixed pairs, or contexts. Each option produces a distribution of outcomes, and participants must learn which option is more valuable in each context by making choices and receiving feedback. Then, in the transfer test, the options are rearranged into new combinations and feedback is no longer provided. To make payoff-maximizing choices, participants need to have learned the absolute reward values of the options during the first phase. However, Hayes and Wedell (2023) found that transfer preferences were systematically biased in favor of the options that were best in their original learning contexts, even when those options were associated with lower absolute payoffs. For example, participants preferred one option over another that yielded $15 more on average, simply because the first option was the best in its original context and the second was not. This result is consistent with the biasing effects of relative valuation; would LLMs exhibit similar behavior?

Here, we show that relative value biases do indeed emerge when LLMs perform a modified version of the bandit task described above. Further, we demonstrate that the effects are not confined to a single model architecture, and they change depending on the structure and content of the prompts. As our ultimate goal is to understand human cognition, we consider the psychological implications of our findings in the Discussion.

## Method

Below we describe the LLMs that we tested, along with our general procedures for collecting responses from the models in the decision-making task.

### Models

We tested four state-of-the-art LLMs: two proprietary models developed by OpenAI (OpenAI, 2023), gpt-4-1106-preview (GPT-4-Turbo) and gpt-3.5-turbo-0613 (GPT-3.5-Turbo), and two open-source models developed by Meta (Touvron et al., 2023), Llama-2-70B and Llama-2-13B. The experiments were run in December, 2023 and January, 2024. All models were optimized for chat-based interactions.

The models were accessed in different ways. We interacted with the GPT models using the Python-based OpenAI API. The 70B parameter version of Llama-2, the latest and biggest model released by Meta, is currently hosted on HuggingChat (v0.7.0), an open-source alternative to ChatGPT.[1] We used the unofficial HuggingChat Python API to automate data collection with Llama-2-70B.[2] We interacted with the quantized 13B parameter version of Llama-2 using the "transformers" package in Python, following the steps outlined in Hussain et al. (2023; see p. 15-16).

LLMs have several parameters that modulate their behavior. We used the default temperature for all models, which allowed for some degree of stochasticity in the responses (as recommended by Demszky et al., 2023). We used a neutral system prompt ("You are a helpful assistant") in the simulations presented here. We also tried the system prompt, "Pretend that you are a human" (Guo, 2023), but it had little effect on the models' behavior (results not shown). All other parameters were kept at their default values.

### Procedure

We implemented a modified version of the bandit task described in Hayes and Wedell (2023). At a high level, the task involves repeated choices between options, each with an initially unknown payoff distribution. The instructed goal is to maximize one's total payoff over the course of several rounds. To collect responses from LLMs, the task must be converted to a text format. We did this by randomly assigning names to each option (e.g., "slot machine G") and describing the outcomes in words (e.g., "delivered 18 dollars"). Second, because LLMs do not have memory for past interactions between the user and model, we followed previous work by including a full history of previous outcomes in each prompt (Binz & Schulz, 2023). Each time the model makes a choice, an outcome is drawn from the corresponding distribution and appended to the history of previous outcomes. The updated history is then included in the prompt for the next trial. Finally, we greatly reduced the number of trials compared to the original version of the task to prevent the prompts from getting too long. To compensate for the smaller number of trials, we reduced the variance of the outcomes so that it would be easier for the models to identify the better options after only a few samples.

There are eight options presented in fixed pairs, or contexts, during the first part of the task (*learning phase*). Feedback is presented from both options after each choice. Due to the structure of the task, each option's outcomes are always presented along with the outcomes of the other option in the same context. The outcomes for each option were drawn from a normal distribution with an option-dependent mean and a standard deviation of 1.0 (Table 1). Outcomes were presented in US dollars and rounded to the nearest integer. The mean outcomes increase in steps of $3 across the eight options. We label each option according to whether it was the lower (L) or higher (H) valued option in its learning context, with a subscript indicating the expected payoff. Henceforth, we refer to options $H_{18}$, $H_{24}$, $H_{30}$, and $H_{36}$ as *locally optimal* and the others as *locally suboptimal*.

**Table 1**: Task structure.

| Context | Low Option | High Option |
|---|---|---|
| Low | $L_{15} \sim N(15, 1)$ | $H_{18} \sim N(18, 1)$ |
| Medium-Low | $L_{21} \sim N(21, 1)$ | $H_{24} \sim N(24, 1)$ |
| Medium-High | $L_{27} \sim N(27, 1)$ | $H_{30} \sim N(30, 1)$ |
| High | $L_{33} \sim N(33, 1)$ | $H_{36} \sim N(36, 1)$ |

The four choice contexts in Table 1 were presented five times each for a total of 20 learning phase trials (randomly ordered).

The second part of the task involves choices between all possible pairs of options without feedback (*transfer test*). With eight options, there are 28 possible pairs, each presented once for a total of 28 transfer test trials (randomly ordered). Crucially, absolute value representations are necessary to choose the payoff-maximizing options in the transfer test. Consider, for example, a choice between $H_{18}$ and $L_{21}$. This combination would not have been previously encountered because the options come from different learning contexts. If

---

[1] https://huggingface.co/chat?model=meta-llama/Llama-2-70b-chat-hf

[2] https://github.com/Soulter/hugging-chat-api

```
You are playing a game with the goal of winning as much money as possible over the course of several rounds.
In each round, you will be asked which of two slot machines you wish to play.
Some slot machines win more money than others on average.
Your total payoff will be the cumulative sum of the money you win across all rounds of the game.
Remember that your goal is to maximize your total payoff.

You made the following observations in the past:

- In Round 1, slot machine B delivered 33 dollars and slot machine C delivered 36 dollars.
- In Round 2, slot machine H delivered 23 dollars and slot machine A delivered 20 dollars.
- In Round 3, slot machine H delivered 25 dollars and slot machine A delivered 20 dollars.
- In Round 4, slot machine G delivered 27 dollars and slot machine D delivered 30 dollars.
- In Round 5, slot machine D delivered 29 dollars and slot machine G delivered 26 dollars.
- In Round 6, slot machine F delivered 14 dollars and slot machine E delivered 18 dollars.
- In Round 7, slot machine C delivered 37 dollars and slot machine B delivered 32 dollars.
- In Round 8, slot machine E delivered 19 dollars and slot machine F delivered 16 dollars.
- In Round 9, slot machine C delivered 34 dollars and slot machine B delivered 32 dollars.
- In Round 10, slot machine F delivered 15 dollars and slot machine E delivered 16 dollars.

You now face a choice between slot machine E and slot machine F.
Your goal is to maximize your total payoff over the course of several rounds.
Which slot machine do you choose? Give your answer without explaining your reasoning.

 I choose slot machine E.
```

**Figure 1**: Example of a prompt and the model's response in the "Baseline" condition (see **Method**).

the decision maker learned the absolute reward values associated with each option, they should correctly choose $L_{21}$ over $H_{18}$. On the other hand, if the decision maker learned only the relative values, they might choose $H_{18}$, which was locally optimal, over $L_{21}$, which was locally suboptimal. Hayes and Wedell (2023) found that participants frequently chose locally optimal options with lower absolute payoffs over locally suboptimal options with higher absolute payoffs, consistent with a relative value bias.

Figure 1 shows an example prompt for a learning phase trial and the model's response in the Baseline condition. A description of the task and the goal ("maximize your total payoff") was prepended to every prompt, followed by a record of the outcomes from previous rounds. The last part of the prompt indicated the current pair of options to choose from and a reminder of the task goal. The models were instructed to indicate their choice without explaining their reasoning. For the transfer test, the prompts were formatted in the same way and included the outcomes from all 20 learning phase trials. Because transfer choices were made without feedback, no additional lines were added to the outcome history after Round 20, the last learning phase trial.

We ran 30 simulations of the bandit task for each model. The eight slot machines were randomly assigned to the letters A through H at the start of every simulation run. On each trial, the presentation order of the two available slot machines was randomized. The order of the options in the record of previous outcomes reflected the order in which they were presented in the choice prompt.

## Results

We begin by examining performance in the learning phase. To be precise, when we say that an LLM successfully "learned" the task, we mean that the model exhibited an increasing tendency to choose the locally optimal options as the number of previous outcomes in the prompt increased (i.e., in-context learning). Simply put, a model should become more accurate when given more information on previous outcomes.

Figure 2 shows the proportion of times the models selected the higher-valued (correct) option on the five learning trials for each context. The proportions were computed across the 30 simulation runs for each model.

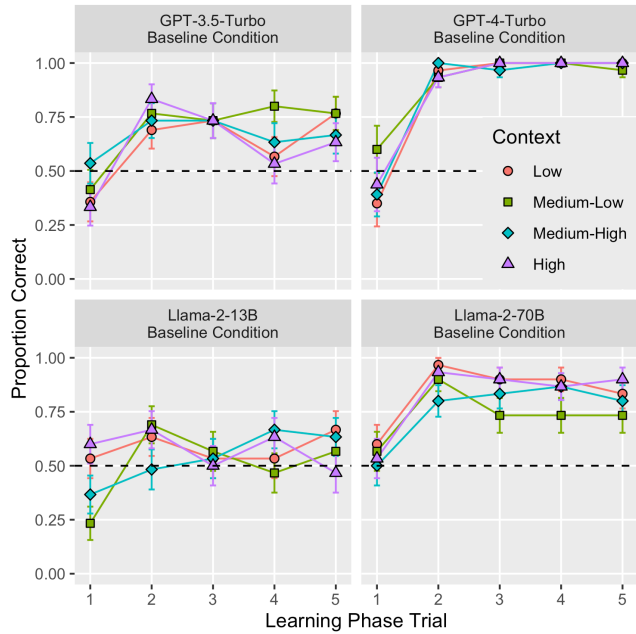

**Figure 2**: Learning phase performance. Each point represents the proportion of times a model chose the locally optimal option from a particular context on a particular trial. Standard error bars are shown.

On the first trial, accuracy was near chance due to the absence of information on previous outcomes.[3] However, as the amount of information increased, two of the four models (GPT-4-Turbo and Llama-2-70B) showed a clear trend of selecting the correct options more frequently. By the fifth and final opportunity, GPT-4-Turbo was choosing the locally optimal option nearly 99% of the time across the four contexts (significantly above chance; $t(29) = 59$, $p < .001$, $d = 10.77$). Llama-2-70B reached a final-trial accuracy around 82% ($t(29) = 7.99$, $p < .001$, $d = 1.46$). In contrast, GPT-3.5-Turbo showed weaker evidence of learning, reaching a final-trial accuracy around 71% ($t(29) = 5.22$, $p < .001$, $d = 0.95$), and Llama-2-13B showed little to no evidence of learning, with a final-trial accuracy around 58% (not significantly different from chance; $t(29) = 1.98$, $p = .06$, $d = 0.36$). Based on these results, we restricted the remaining analyses to the two most successful models: GPT-4-Turbo and Llama-2-70B.

The transfer test presented choices between all possible pairs of options without feedback. If the models rely on absolute value representations, they should be unaffected by the novel option pairings and highly accurate in choosing the payoff-maximizing options. In contrast, if the models rely on relative value representations, they should exhibit a bias toward selecting the options that were locally optimal in their original contexts. One way to tell is by examining choice rates: the number of times an option was chosen divided by the number of times it was presented (each option appeared 7 times in the transfer test). Pure absolute valuation produces choice rates that increase monotonically and linearly across the eight options, while pure relative valuation produces a "zig-zag" pattern driven by frequent choices of the options that were locally optimal in the learning phase (Figure 3).

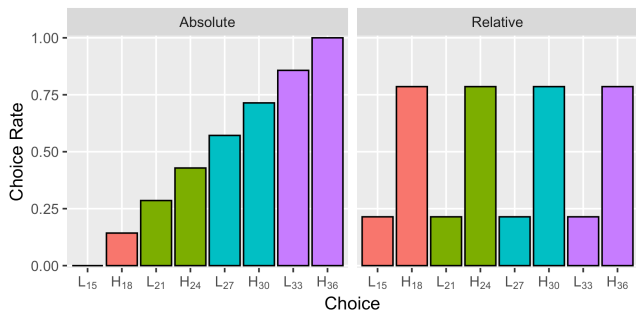

**Figure 3**: Theoretical transfer choice rates under absolute versus relative valuation.

Figure 3 suggests a quick and easy way to test for absolute versus relative valuation: The average choice rate for options $L_{21}$, $L_{27}$, and $L_{33}$ should be greater than the average choice rate for options $H_{18}$, $H_{24}$, and $H_{30}$ under absolute valuation, but the reverse should be true under relative valuation. Thus, we compute a linear contrast on the choice rates using the following coefficients for the eight options, $L_{15}$ through $H_{36}$: [0, –1/3, 1/3, –1/3, 1/3, –1/3, 1/3, 0]. If the mean value of this contrast is significantly positive, it supports absolute valuation; if the mean value is significantly negative, it supports relative valuation.

Figure 4 shows the transfer choice rates for GPT-4-Turbo (top) and Llama-2-70B (bottom), with the Baseline condition in the leftmost panels. A clear zig-zag pattern is evident for GPT-4-Turbo, consistent with relative value bias, and the mean contrast score was both significant and negative. The pattern for Llama-2-70B was noisier overall with some evidence of a weak bias, but the mean contrast did not significantly differ from zero. As shown in Table 2, GPT-4-Turbo tended to select lower-valued options that were locally optimal in their original contexts over higher-valued options that were locally suboptimal, but only when the difference in expected payoffs was small ($3). Llama-2-70B chose the correct option at near-chance levels for these choice pairs.

The remaining panels in Figure 4 show the results from additional experimental conditions. The first two conditions were adapted from Hayes and Wedell (2023). In the *Feelings* condition, the last part of each prompt included instructions to "think about how positive or negative you currently feel about slot machine [] and slot machine []", and to "rate your feelings for both slot machines using an integer from 1 to 7, where 1 = Extremely negative and 7 = Extremely positive." In the *Expected Outcomes* condition, the instructions were instead to "think about how much money you would expect to win from slot machine [] and slot machine []," and to "estimate the amount of money (in dollars) you would expect to win if you chose each slot machine." In both cases, the first five lines of the prompt and the outcome history were the same as in the Baseline condition. Analogous to human participants (Hayes & Wedell, 2023), relative value biases were greatly attenuated in both GPT-4-Turbo and Llama-2-70B when the models were instructed to think about expected outcomes. Interestingly, the models' choices aligned almost perfectly with absolute values in this condition, which was not the case for the participants in that study. Instructing the models to think about their feelings for each option had little effect compared to baseline. The feeling ratings themselves, however, clearly exhibited the characteristic zig-zag pattern of relative valuation (results not shown here).

Two other novel conditions were tested. In the *Regret* condition, each line of the outcome history was worded as follows: "In Round [], slot machine [] delivered [] dollars, which is [] dollars [more / less] than slot machine [] delivered ([] dollars)." For example, if slot machine B delivered 19 dollars and slot machine E delivered

---

[3] Models occasionally gave an ambiguous response or refused to choose. This was especially common on the first trial for a context. Such trials were removed prior to analysis (2.9% of the responses for GPT-4-Turbo, 0.3% for Llama-2-13B, 1% for GPT-3.5-Turbo).

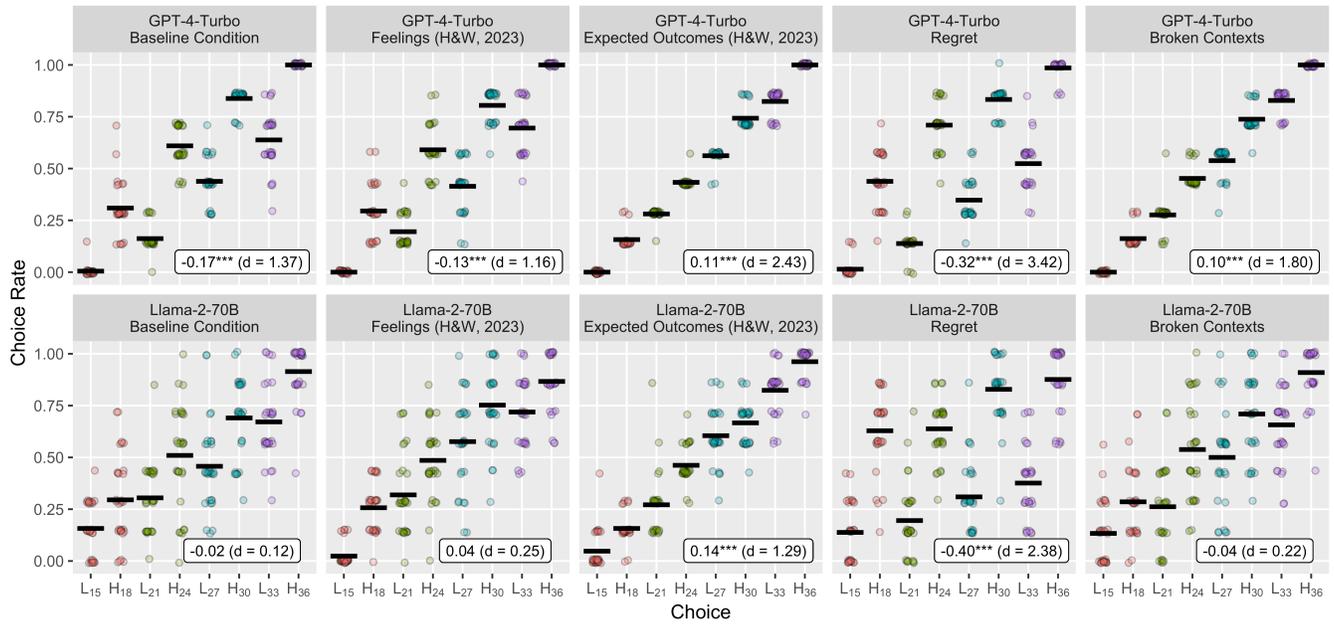

**Figure 4**: Transfer test choice rates. Each point represents the choice rate for a particular option on a single simulation run. Bars show the means across simulation runs. A linear contrast was computed by subtracting the mean of the choice rates for $H_{18}$, $H_{24}$, and $H_{30}$ from the mean of the choice rates for $L_{21}$, $L_{27}$, and $L_{33}$. Positive values are consistent with absolute valuation, and negative values are consistent with relative valuation. The bottom right of each panel shows the average value of this contrast, the results of a significance test (one-sample t-test, $n = 30$, $H_0: \mu = 0$), and the effect size (Cohen's $d$). ***$p < .001$.

15 dollars in Round 5, the last line of the outcome history on Round 6 would read, "In Round 5, slot machine B delivered 19 dollars, which is 4 dollars more than slot machine E delivered (15 dollars)." The instructions in the prompt were left unchanged. We hypothesized that the presence of explicit outcome comparisons would magnify relative value biases. Consistent with our hypothesis, both models showed large, highly significant relative value biases in this condition, more so than in any other (Figure 4). Llama-2-70B also exhibited a striking preference for option $H_{18}$ over $L_{33}$, choosing $L_{33}$ on just 6 out of 30 runs, despite its $15 advantage in expected payoffs (Table 2).

In the *Broken Contexts* condition, we broke up the outcome contexts by listing each outcome on a separate line and removing the numbers of the rounds. Outcomes were still listed in the order in which they were observed. For the previous example, the last *two* lines of the outcome history on Round 6 would read, "slot machine B delivered 19 dollars," followed by, "slot machine E delivered 15 dollars." The instructions were left unchanged. We predicted diminished relative value biases in this condition due to the lack of explicit contextual structure in the outcome history. Our prediction was confirmed for GPT-4-Turbo, which exhibited little to no bias, but the manipulation appeared to have little effect for Llama-2-70B.

In summary, both models showed signs of relative value bias in the baseline condition, especially GPT-4-Turbo. The bias was eliminated by prompting the models to think about (and estimate) expected outcomes prior to making a choice. The bias was magnified when the differences between outcomes in each context were explicitly stated in the prompt. Finally, for GPT-4-Turbo, breaking the contextual structure of the outcome history by listing each outcome on a separate line all but eliminated relative value biases.

## Discussion

The present study demonstrates that two of the highest performing LLMs currently available, GPT-4-Turbo and Llama-2-70B, exhibit relative value decision biases analogous to those observed in prior human and animal studies (e.g., Bavard et al., 2018; Hayes & Wedell, 2023; Pompilio & Kacelnik, 2010; see Palminteri & Lebreton, 2021, for a review). This finding is remarkable for a number of reasons, which we outline below.

Every prompt that was fed to the models contained a complete record of previous outcomes from both options. Therefore, in the transfer test, the models had simultaneous access to all of the necessary information to "figure out" which option had produced higher absolute payoffs in the past. Despite this, the models' choices were still affected by relative values. This result is in some ways similar to the finding that LLMs exhibit properties of memory qualitatively similar to those observed in human memory studies, such as primary and recency effects, even though the to-be-recalled information is directly available in the prompt (Janik, 2023). Evidently, just because a model has access to the necessary information, it does not necessarily mean that it will utilize the information appropriately.

Table 2: Payoff-maximizing choice proportions for diagnostic choice pairs in the transfer test.

| Model | Condition | $H_{18}$ vs. $L_{21}$ | $H_{24}$ vs. $L_{27}$ | $H_{30}$ vs. $L_{33}$ | $H_{18}$ vs. $L_{27}$ | $H_{24}$ vs. $L_{33}$ | $H_{18}$ vs. $L_{33}$ |
|---|---|---|---|---|---|---|---|
| GPT-4-Turbo | Baseline | **.17\*\*\*** | **.13\*\*\*** | **.13\*\*\*** | .80\*\* | .57 | .90\*\*\* |
| | Feelings | **.30\*** | **.20\*\*** | **.27\*** | .70\* | .69\* | 1.00\*\*\* |
| | Exp. Outcomes | .97\*\*\* | 1.00\*\*\* | .80\*\* | .93\*\*\* | .97\*\*\* | 1.00\*\*\* |
| | Regret | **.03\*\*\*** | **.03\*\*\*** | **.03\*\*\*** | **.31\*** | **.14\*\*\*** | .63 |
| | Broken Contexts | .87\*\*\* | .80\*\* | .80\*\* | 1.00\*\*\* | 1.00\*\*\* | 1.00\*\*\* |
| Llama-2-70B | Baseline | .50 | .53 | .47 | .63 | .57 | .87\*\*\* |
| | Feelings | .60 | .63 | .57 | .87\*\*\* | .67 | 1.00\*\*\* |
| | Exp. Outcomes | .80\*\* | .80\*\* | .93\*\*\* | 1.00\*\*\* | .93\*\*\* | .97\*\*\* |
| | Regret | **.13\*\*\*** | **.17\*\*\*** | **.10\*\*\*** | **.17\*\*\*** | **.17\*\*\*** | **.20\*\*** |
| | Broken Contexts | .50 | .53 | .43 | .80\*\* | .60 | .87\*\*\* |

*Note*: These choice pairs pit an option with lower payoffs that was locally optimal in its original context (H) against an option with higher payoffs that was locally suboptimal in its original context (L). The payoff-maximizing choice for these pairs is L. Choice proportions significantly above chance are consistent with absolute valuation. Choice proportions significantly below chance (bold) are consistent with relative valuation. \*$p < .05$ \*\*$p < .01$ \*\*\*$p < .001$ (z-test for a proportion).

The models' behavior changed in response to experimental manipulations in ways that resembled human behavior. Instructing the models to think about and estimate expected outcomes prior to making a choice all but eliminated relative value biases. A similar manipulation reduced, but did not eliminate, relative value bias in humans (Hayes & Wedell, 2023). We also found that we could magnify the bias by adding explicit, contextual outcome comparisons to the outcome history. In human studies (Bavard et al., 2018, 2021), relative value biases are enhanced when participants are given the outcomes from both options (complete feedback), as opposed to just the outcomes of chosen options (partial feedback), presumably due to the greater ease of making such counterfactual comparisons. Complete feedback was always available in the experiments presented here, but future studies could test whether relative value biases can be further enhanced in human participants by emphasizing the *differences* between outcomes on each feedback presentation.

The choice biases that we observed may have other contributing sources beyond relative valuation in humans. However, some of these factors can be quickly ruled out in the case of LLMs. For example, because LLMs do not have intrinsic goals, we can rule out the possibility that relative value biases in these models are driven by intrinsic signals of goal achievement (Molinaro & Collins, 2023). Second, relative values are often confounded with choice history in human experiments: The options with better relative values also tend to be chosen more frequently in the learning phase, and therefore may be selected over options with higher absolute payoffs in the transfer test simply out of habit, or the tendency to repeat previous actions (Miller, Shenhav, & Ludvig, 2019; see Bavard et al., 2021, for a discussion). Here, the LLMs were not given any information about their previous choices. Thus, we can rule out habit formation as a potential explanation. Every trial is a completely independent interaction with the LLM; there is no way for the models to develop a propensity for repeating previous choices.

Our results instead suggest that relative value biases can emerge solely from the nature of language processing in LLMs. There are at least two possible reasons why human-like characteristics might emerge from the behavior of LLMs, and they are not mutually exclusive. One is that the characteristics arise due to specific features of the model architecture or its inner workings. The other possibility is that the characteristics reflect statistical correlations or patterns that are present in the models' training data. To paraphrase Janik (2023), if we as humans structure our texts in ways that are consistent with our own cognitive characteristics (including our biases), then LLMs, which are essentially massive statistical models trained to detect patterns in text data, might end up learning and exhibiting these characteristics.

The fact that we demonstrated relative value biases in both GPT-4-Turbo and Llama-2-70B suggests that the effects are not due to specific features of either model. However, future studies should examine a wider range of models with different architectures, training datasets, and finetuning methods to better assess the generalizability of our findings.

Future studies might also consider looking "under the hood," examining the internals of the models to better understand the emergent computational mechanisms that give rise to relative value bias. In the case of decoder models like GPT and Llama-2, which are trained to predict the next token in a sequence, the contextualized embedding for each token depends on all of the tokens preceding it (Hussain et al., 2023). Thus, by examining the embeddings associated with different options (or outcomes) at different points in the learning phase, it might be possible to visualize the emergence of relative valuation in LLMs. One might expect the representations for the locally optimal options in each context to become gradually more similar to each other, and less similar to the representations for the locally suboptimal options, as the learning phase progresses. Such an approach could also be used to understand how different experimental manipulations affect the models' underlying representations of the task.